\documentclass[]{interact}

\usepackage{epstopdf}
\usepackage[caption=false]{subfig}  
\usepackage[numbers,sort&compress]{natbib}
\bibpunct[, ]{[}{]}{,}{n}{,}{,}

\theoremstyle{plain}
\newtheorem{theorem}{Theorem}[section]

\theoremstyle{definition}
\newtheorem{definition}[theorem]{Definition}

\theoremstyle{remark}

\usepackage{amsmath, amssymb, amsthm}
\usepackage{mathtools}
\usepackage{amsfonts}
\usepackage{bm}
\usepackage{xcolor}
\usepackage{geometry}
\usepackage{graphicx}
\usepackage{float}
\usepackage{booktabs}
\usepackage{algorithmic}
\usepackage[ruled,vlined]{algorithm2e}
\usepackage{tikz}
\usetikzlibrary{shapes.geometric, arrows.meta, arrows, positioning, fit, backgrounds}
\usepackage{caption}          
\begin{document}

\articletype{ARTICLE TEMPLATE}

\title{Bridging Computational Social Science and Deep Learning: Cultural Dissemination-Inspired Graph Neural Networks}

\author{
\name{Asela Hevapathige\textsuperscript{a}\thanks{CONTACT Asela Hevapathige. Email: asela.hevapathige@anu.edu.au}}
\affil{\textsuperscript{a}School of Computing, Australian National University, Canberra, Australia}
}

\maketitle

\begin{abstract}
Graph Neural Networks (GNNs) have become vital in applications like document classification in citation networks, epidemic forecasting, viral marketing, user recommendation in social networks, and network monitoring.  However, their deployment faces three key challenges: feature oversmoothing in deep architectures, poor handling of heterogeneous relationships, and monolithic feature aggregation. To address these, we introduce AxelGNN, a novel architecture based on Axelrod's 
cultural dissemination model that incorporates three key innovations: (1) 
similarity-gated interactions that adaptively promote convergence or divergence 
based on feature similarity, (2) segment-wise feature copying that enables 
fine-grained aggregation of semantic feature groups rather than monolithic 
vectors, and (3) global polarization that maintains multiple distinct 
representation clusters to prevent oversmoothing. This model demonstrates empirically the capability to handle both homophilic and 
heterophilic graphs within a single architecture, without requiring specialized 
model selection based on graph characteristics. Our experiments demonstrate that AxelGNN achieves competitive or superior performance 
compared to existing methods in node classification and influence estimation while 
maintaining computational efficiency.
\end{abstract}


\section{Introduction}

Graph Neural Networks (GNNs) have been a cornerstone to many graph downstream tasks, such as node classification \cite{xiao2022graph,liang2022survey}, influence maximisation \cite{li2023survey,jaouadi2024survey}, link prediction \cite{zhang2018link,arrar2024comprehensive}, and community detection \cite{souravlas2021survey,su2022comprehensive}.  These capabilities of GNNs translate to many practical applications, including document categorization in citation networks, epidemic intervention planning, recommendation systems, protein function prediction, and social network analysis. This success can be attributed to their ability to effectively leverage the structural relationships within graph data, which enables the learning of rich node representations that seamlessly integrate both node-level features and topological information through iterative message passing mechanisms \cite{gilmer2017neural}. Despite their widespread adoption, practitioners face three critical limitations when deploying GNNs in real-world applications:

\begin{itemize}
    \item \textbf{Feature Oversmoothing Degrades Deep Models:} As GNNs stack more layers, node representations become increasingly similar to their neighbours, eventually converging to nearly identical embeddings across the entire graph \cite{rusch2023survey}. This reduces the model's ability to distinguish between different nodes and hurts performance, especially in downstream tasks that require deeper networks to capture long-range dependencies, such as epidemic spread prediction where multi-hop transmission chains must be modeled or citation network node classification where categories depend on multi-hop  relationships.\\
    
    \item \textbf{Poor Performance on Heterogeneous Real-World Networks:} Traditional GNNs assume that connected nodes tend to have similar features and labels, known as the homophily assumption \cite{zhu2020beyond}. However, many real-world graphs exhibit heterophily, where connected nodes are dissimilar \cite{luan2022revisiting}. In web page networks, pages frequently link across different categories bridging diverse content types. Similarly, epidemic contact networks are inherently heterophilic, as disease transmission occurs between infected and susceptible individuals with different states. GNNs perform poorly on such heterophilic graphs because their message-passing mechanism reinforces similarity between neighbours, which contradicts the underlying graph structure.\\
    
    \item \textbf{Monolithic Feature Treatment:} Existing GNNs treat node features as indivisible units during aggregation. The entire feature vector is processed as a single entity, without considering the individual importance or relationships between different feature dimensions. This prevents fine-grained feature-level interactions and limits the model's ability to selectively aggregate specific aspects of neighboring node features.
\end{itemize}

We seek to address the limitations above. We observe that Axelrod's cultural dissemination model \cite{axelrod1997dissemination} exhibits remarkable properties that directly correspond to the fundamental challenges in GNNs. The model's dual convergence behavior naturally handles both homophilic and heterophilic relationships: similarity convergence (where neighboring agents become identical) mirrors beneficial feature sharing in homophilic graphs, while dissimilarity convergence (where neighboring agents become completely different) preserves node distinctiveness in heterophilic scenarios. Additionally, the model's global polarization mechanism prevents the collapse to a single representation by maintaining multiple distinct cultural regions, directly addressing the oversmoothing problem. Finally, Axelrod's trait-level copying mechanism operates on individual cultural dimensions rather than entire cultural vectors, providing a principled approach to fine-grained feature interactions that transcends monolithic feature treatment.

Motivated by these observations, we propose \emph{AxelGNN}, a novel graph neural network architecture that incorporates the core principles of Axelrod's cultural dissemination model into the neural message-passing framework. Our approach introduces similarity-gated interactions that adaptively promote convergence or divergence based on node similarity, implements trait-level copying mechanisms for fine-grained feature aggregation, and maintains global polarization to preserve node distinctiveness across multiple representation clusters. Our contributions advance the practical deployment of GNNs as follows:

\begin{itemize}
    \item \textbf{Unified Framework for Diverse Graph Types:} We introduce a single architecture that effectively handles both homophilic and heterophilic graphs, eliminating the need for practitioners to select specialized models based on graph characteristics---a practical advantage for real-world deployments where graph properties may be unknown or mixed.\\
    
    \item \textbf{Novel GNN Architecture:} We propose AxelGNN, which maintains performance across varying network depths without oversmoothing degradation, enabling practitioners to build deeper models for applications requiring multi-hop reasoning, such as node classification and epidemic forecasting.\\
    
    \item \textbf{Comprehensive Empirical Validation:} We conduct extensive experiments across diverse real-world applications, including node classification on citation networks and social graphs, and influence estimation on epidemic diffusion networks, demonstrating superior or competitive performance compared to existing methods.\\
    
    \item \textbf{Practical Implementation Guidance:} We provide efficiency analysis, hyperparameter sensitivity studies, and a simplified variant that offers practitioners trade-offs between model expressiveness and computational requirements based on their application constraints.
\end{itemize}

\section{Related Work}

\subsection{Traditional GNNs} 

Graph Neural Networks have been increasingly used for diverse graph-related downstream tasks \cite{wu2020comprehensive}. There have been many GNN types proposed with different inductive biases. Gilmer et al. \cite{gilmer2017neural} proposed one of the early GNNs based on the message-passing paradigm. Graph Convolutional Networks (GCNs) introduced a spectral-based perspective where convolution operations are defined in the spectral domain of graphs using the graph Laplacian matrix \cite{kipf2017semi}. Graph Attention Networks (GATs) consider a learnable neighbourhood weighting mechanism for message-passing, going beyond traditional fixed, structure-based weighting schemes \cite{velivckovic2018graph}. GraphSAGE is another GNN designed based on a sampling mechanism to enhance GNN's scalability to large graphs \cite{hamilton2017inductive}. However, these classical GNNs are known to work under the homophily assumption, thus cannot model heterogeneous relationships among neighbouring nodes, and are prone to oversmoothing. 

\subsection{Heterophily-aware GNNs} 
Pei et al\cite{pei2020geom} presented one of the early works that identified the performance degradation issue in GNNs under heterophilic data. Since then, researchers have proposed GNNs enhanced by various techniques to alleviate this limitation. Some of these works have proposed neighbourhood filtering to aggregate information from relevant neighbours selectively \cite{bo2021beyond,luan2022revisiting,guo2024gnn}. In contrast, some other works have introduced higher-order graph structures to capture multi-hop relationships that may be more informative than direct connections \cite{zhu2020beyond,wang2021graph,wang2022powerful,li2024pc}. There have been works utilising specific design decisions, such as tailored aggregation mechanisms \cite{he2022block,haruta2023novel} and neighbourhood ordering \cite{songordered}, to enhance relevant information flow. These approaches aim to suppress noise flow in message passing. Recent advances include FAGCN~\cite{bo2021beyond}, which employs adaptive frequency-based filtering to combine low-pass and high-pass graph signals, ES-GNN~\cite{guo2024gnn}, which addresses heterophily through edge splitting that separates homophilic and heterophilic connections, and AD-GNN \cite{hevapathige2025beyond} that adaptively change the aggregation depth to accomodate node-level homophily. However, certain limitations still exist. First, many existing approaches cannot handle both homophilic and heterophilic data patterns within a single framework effectively. Most methods are specifically designed for either homophilic or heterophilic scenarios, limiting their practical applicability in real-world graphs that often exhibit mixed characteristics. These methods typically require explicit signal processing, graph restructuring, or specialized aggregation schemes tailored to specific homophily levels. In contrast, AxelGNN addresses heterophily through bistable convergence dynamics inspired by Axelrod's cultural dissemination model, where the same similarity-gated mechanism naturally handles both homophily and heterophily adaptively without requiring frequency analysis or structural modifications. Second, existing methods often lack the flexibility to adaptively determine when to emphasise local versus global information based on neighbourhood characteristics, resulting in suboptimal performance when the optimal aggregation strategy varies across different parts of the same graph. 

\subsection{Oversmoothing in GNNs} 
Oversmoothing in GNNs refers to the scenario where node features become increasingly similar and eventually indistinguishable in the embedding space as message passing iterations increase \cite{rusch2023survey}. Existing works try to alleviate this issue through various mechanisms. Some works propose structural modifications, such as residual connections \cite{yin2022adaptive,yang2022difference,scholkemperresidual} and jumping knowledge mechanisms \cite{yang2022graph,zhu2023matrix}, to retain initial representations of nodes to some extent, thus preventing the complete loss of node distinctiveness. Some other works introduced regularisation and normalisation techniques, such as dropout \cite{luo2021learning,shu2022understanding,hssayni2025novel,boufssasse2025addressing} and batch normalisation techniques \cite{zhaopairnorm,zhou2020towards}, to introduce controlled stochasticity and maintain feature diversity throughout the graph. Recently, researchers have used curvature-based methods to identify bottlenecks in graphs that lead to oversmoothing \cite{nguyen2023revisiting,liu2023curvdrop,fesser2024mitigating,hevapathige2025depth}. They also proposed graph rewiring and sampling methods to structurally correct these bottlenecks. However, most existing approaches still suffer from several limitations. First, they typically apply uniform treatments across the entire graph without considering local neighborhood characteristics, potentially over-correcting in regions where information sharing would be beneficial while under-correcting in areas where distinctiveness preservation is crucial. For instance, dropout-based methods apply stochastic masking uniformly, while curvature-based approaches require expensive geometric computations and graph modifications. AxelGNN fundamentally differs by preventing oversmoothing through global polarization: the similarity-gated interaction mechanism creates multiple distinct representation clusters where nodes within clusters can converge while nodes across clusters maintain separation. This polarization emerges naturally from cultural dissemination dynamics without requiring stochastic dropout or curvature computation, and adapts to local graph structure which promotes information flow in homophilic regions while preserving distinctiveness in heterophilic regions. Second, many methods lack principled theoretical foundations for determining when and how to prevent oversmoothing, often relying on heuristic solutions that may not generalise across different graph types and diverse downstream tasks. Finally, these approaches often fail to provide adaptive mechanisms that can simultaneously handle both homophilic regions where controlled smoothing is desirable and heterophilic regions where preservation of differences is essential within the same graph structure.

\subsection{Novelty of Our work}

AxelGNN introduces a novel approach that differs from existing heterophily 
and oversmoothing GNNs by providing a unified framework grounded in Axelrod's 
cultural dissemination theory. While existing methods typically target either 
homophilic or heterophilic scenarios through specialized architectures, AxelGNN 
demonstrates through empirical evaluation that a single architecture inspired 
by cultural dissemination dynamics can adapt to diverse graph characteristics. The method's trait copying mechanism breaks the monolithic feature assumption by operating on grouped feature segments rather than entire feature vectors. This enables fine-grained modeling of complex relationships where different feature dimensions may exhibit varying similarity patterns. This approach naturally prevents oversmoothing through global polarization while maintaining selective information flow, addressing fundamental limitations of existing methods that apply uniform treatments across diverse graph regions.

\section{Background and Motivation}

In this section, we present Axelrod's model of cultural dissemination and discuss its implications for addressing the gaps in GNNs.

\subsection{Axelrod's Cultural Dissemination Model}

The model simulates cultural dissemination on an $L \times L$ grid, where each cell represents an agent with a cultural vector. Each agent $i$ has a cultural vector $\mathbf{v}^i = (v_1^i, v_2^i, \ldots, v_f^i)$ with $f$ cultural dimensions (i.e., traits), where each trait $v_j^i$ can take $q$ possible discrete values from $\{0, 1, \ldots, q-1\}$. Agents start with random trait assignments.

The cultural similarity between two agents $i$ and $j$ is defined as:
\begin{equation}
s_{ij} = \frac{|\{k : v_k^i = v_k^j\}|}{f} 
\end{equation}

where $s_{ij} \in [0,1]$ represents the fraction of matching traits between agents $i$ and $j$.

At each time step, an active agent $k$ and a passive neighbour $r$ are selected randomly. They interact with probability $s_{kr}$, where $s_{kr}$ is their cultural similarity. If interacting, $k$ copies one of $r$'s differing traits at random. The system reaches equilibrium when all neighbouring pairs are either identical ($s_{kr} = 1$) or completely dissimilar ($s_{kr} = 0$). 

\subsection{Convergence and Polarisation in Axelrod Model}

The Axelrod model exhibits two interesting properties in terms of convergence. The model achieves local convergence among neighbours while having global polarisation. Below, we formally define these concepts and explain how the Axelrod model achieves them.

\subsubsection{Local Convergence}

We first define the local convergence property of the Axelrod model, which exhibits bistable dynamics. In this model, neighbouring agents cannot maintain intermediate levels of similarity. Instead, they converge either to complete cultural alignment or to total separation.

\begin{definition}[Local Convergence]
\label{def:local_convergence}
For neighboring agents $i$ and $j$, local convergence occurs when $\lim_{t \to \infty} s_{ij}(t) \in \{0, 1\}$, encompassing:
\begin{itemize}
    \item Similarity convergence: $s_{ij} \to 1$ (identical cultures)
    \item Dissimilarity convergence: $s_{ij} \to 0$ (disjoint cultures)
\end{itemize}
\end{definition}

The bistable behaviour of the Axelrod model emerges from the similarity-dependent interaction mechanism, where agents with higher similarity interact more frequently and subsequently become more alike, while agents with lower similarity interact less frequently and eventually lose all common traits through competing influences from their respective neighbourhoods.

\subsubsection{Global Polarization}

We define the global polarization property of the Axelrod model as a critical equilibrium state in which the system divides into distinct cultural regions. These regions maintain a permanent separation due to the absence of shared traits between neighboring groups.

\begin{definition}[Global Polarization]
\label{def:global_polarization}
Global polarization occurs when the system partitions into $K \geq 2$ cultural groups $\{G_1, \ldots, G_K\}$ where each $G_k$ is a non-empty subset of agents for $k = 1, \ldots, K$, such that:
$$s_{ij} = 0 \quad \forall i \in G_k, j \in G_m, k \neq m, \text{ where } i \text{ and } j \text{ are neighboring agents}$$
\end{definition}

Global polarization of the Axelrod model emerges through progressive boundary sharpening. As local regions homogenize, they lose the trait overlap that enables cross-regional interaction. When $s_{ij} = 0$ between neighboring regions, interaction probability drops to zero, creating permanent cultural boundaries. The same similarity-based mechanism that unifies local neighborhoods simultaneously destroys the connectivity necessary for global consensus, fragmenting the system into isolated cultural domains.

\subsection{Motivation: Addressing Homophily, Heterophily, and Oversmoothing in GNNs}

The Axelrod model's convergence properties provide a principled motivation for 
addressing fundamental challenges in graph neural networks across diverse 
downstream tasks. While not a direct mathematical equivalence, the analogy 
guides our architectural design choices and provides intuitive justification 
for the mechanisms we introduce.

\subsubsection{Mapping to Homophily/ Heterophily Paradigms}
The dual local convergence states in the Axelrod model naturally correspond to homophily-heterophily downstream task scenarios:
\begin{itemize}
    \item Homophily $\leftrightarrow$ Local Similarity Convergence: In homophilic graphs where similar nodes are connected, the model promotes $s_{ij} \to 1$, enabling beneficial feature sharing and alignment between neighboring nodes of the same class
    \item Heterophily $\leftrightarrow$ Local Dissimilarity Convergence: In heterophilic graphs where dissimilar nodes are connected, the model drives $s_{ij} \to 0$, preserving node distinctiveness and preventing harmful feature averaging between nodes of different classes
\end{itemize}

\subsubsection{Global Polarization Prevents Oversmoothing}

Traditional GNNs suffer from oversmoothing because all nodes converge to a single representation as depth increases \cite{zhang2023comprehensive}. The Axelrod model's global polarization mechanism directly addresses this fundamental limitation. Rather than converging to a single global minimum, the system fragments into multiple cultural regions, naturally preserving multiple distinct node representations. The formation of stable cultural boundaries with $s_{ij}=0$ between regions prevents different node classes from collapsing into identical representations. Furthermore, similarity-gated interactions ensure that information flows within semantically coherent regions while maintaining separation between different classes, enabling selective information flow that preserves node distinctiveness throughout the learning process.

\section{Methodology}

In this section, we propose a novel GNN architecture called AxelGNN based on the Axelrod cultural dissemination model. We start by providing preliminaries and then offer a comprehensive explanation of each component.

\subsection{Preliminaries}

Let $G = (V, E)$ be an undirected graph where $V$ and $E$ are the node and edge sets, respectively. Each node $v \in V$ has neighborhood $N(v) = \{u | (u, v) \in E\}$. Further, $x_v \in \mathbb{R}^d$ represents initial node features of node $v$ with the feature dimension of $d$. In each GNN layer, the node features would represent a transformation with respect to the node features in the previous layer and the graph structure. We denote the node feature of $v$ in layer $l$ to be $x_v^l \in \mathbb{R}^{d^l}$ where $d^{(l)}$ is the feature dimensionality at layer $l$. 
Throughout this paper, we use superscripts in parentheses (i.e., $x_v^{(l)}$, $h_u^{(l)}$) to denote layer indices, and subscripts (i.e., $x_v$, $h_u$) to denote node indices. Feature dimensions are denoted by $d$ with appropriate layer superscripts (i.e., $d^{(l)}$).

\subsection{Interaction Probability Computation}

We start by computing the interaction probability for each neighbouring node pair. For layer $l$, node features are first transformed via a learnable function: 

\begin{equation}
\mathbf{h}_u^{(l)} = \mathbf{W}^{(l)} \mathbf{x}_u^{(l-1)} + \mathbf{b}^{(l)}   
\end{equation}

where $W^{(l)} \in \mathbb{R}^{d^{(l-1)} \times d^{(l)}}$ is the weight matrix, and $b^{(l)} \in \mathbb{R}^{d^{(l)}}$ is the bias vector with learnable parameters. Here, $l$ denotes the layer index consistently throughout the model architecture. Feature similarity between nodes $v$ and $u$ is computed as:
\begin{equation}
s_{vu}^{(l)} = \frac{\mathbf{h}_v^{(l)} \cdot \mathbf{h}_u^{(l)}}{\|\mathbf{h}_v^{(l)}\|_2 \|\mathbf{h}_u^{(l)}\|_2 + \epsilon}
\end{equation}

Note that $\epsilon$ is a small constant employed for numerical stability (i.e., to avoid division by zero if either vector has zero norm). The interaction probability between nodes $v$ and $u$ at layer $l$ is computed as:
\begin{equation}
p_{vu}^{(l)} = f(s_{vu}^{(l)})
\end{equation}
where $f: [-1,1] \to [0,1]$ is a monotonically increasing interaction function that maps feature similarity to interaction probability. We implement this interaction function using a sigmoid ($\sigma$) formulation that provides interpretable and learnable control over interaction dynamics:

\begin{equation}
f(s) = \sigma(\beta \cdot (s - \theta))
\end{equation}

Then, the complete interaction probability becomes:

\begin{equation}
p_{vu}^{(l)} = f(s_{vu}^{(l)}) = \sigma(\beta \cdot (s_{vu}^{(l)} - \theta))
\end{equation}

where $\beta$ is the intensity parameter and $\theta$ is threshold parameter . This functional form was chosen for several important reasons. The sigmoid function naturally provides bounded output in the range $(0,1)$, ensuring valid probability values, while maintaining monotonic behavior that preserves similarity ordering. The differentiability of the sigmoid function throughout its domain enables effective gradient-based learning during neural network training.  The threshold parameter $\theta \in [-1,1]$ determines the minimum similarity required for meaningful interaction. Higher values establish restrictive thresholds suitable for heterophilic graphs, while lower values create permissive thresholds that accommodate homophilic structures. The intensity parameter $\beta \in \mathbb{R}^+$ controls the steepness of the sigmoid transition, regulating the trade-off between information flow and node diversity preservation. Small values promote broad information exchange but risk oversmoothing, while large values create selective interactions that maintain node distinctiveness.

\subsection{Message Passing and Aggregation}

Feature influence from neighbor $u$ to node $v$ at layer $l$ is computed as:
\begin{equation}
m_{u \to v}^l = p_{uv}^l \cdot h_u^l
\end{equation}

Node $v$  aggregates messages from its neighborhood $\mathcal{N}(v)$:
\begin{equation}
a_v^l = \frac{1}{|\mathcal{N}(v)|} \sum_{u \in \mathcal{N}(v) \cup \{v\}} m_{u \to v}^l
\end{equation}

Next, we implement trait copying by grouping continuous features into segments of size $s$, mirroring Axelrod's discrete cultural traits. In Axelrod's model, traits are adopted as complete units, not fractionally. Similarly, node features contain semantic groups that should be updated together. The segment size $s$ controls this granularity. Let $C = \lceil d^{(l)}/s \rceil$ be the number of groups. For each group $j$, a copying probability is computed using a neural network $\phi_j : \mathbb{R}^{2s} \to \mathbb{R}^s$:

\begin{equation}
\mathbf{c}_{v,j} = \phi_j([\mathbf{h}_v^l[js:\min((j+1)s, d^{(l)})], \mathbf{a}_v^l[js:\min((j+1)s, d^{(l)})]])
\end{equation} 

Here, $[js:\min((j+1)s, d^{(l)})]$ denotes the feature slice corresponding to 
group $j$, where $j$ indexes from $0$ to $C-1$, and $s$ is the segment/group size 
hyperparameter. The updated node representation combines the original and aggregated values for each group:
\begin{align}
\mathbf{x}_v^l[js\!:\!\min((j+1)s, d^{(l)})] &= \mathbf{c}_{v,j}^l \odot \mathbf{a}_v^l[js\!:\!\min((j+1)s, d^{(l)})] \notag \\
&\quad + (\mathbf{1} - \mathbf{c}_{v,j}^l) \odot \mathbf{h}_v^l[js\!:\!\min((j+1)s, d^{(l)})]
\end{align}

where $\odot$ denotes element-wise multiplication and the group size $s$ controls the granularity of trait copying.

\subsection{A Simplified Variant}

In this section, we present a lightweight variant of AxelGNN called AxelGNN$_{\text{Sim}}$. This variant replaces the group-specific neural networks from Eq. 6 with simple learnable weights, reducing the complexity of AxelGNN by a significant margin. 

Instead of computing copying probabilities using group-specific neural networks $\phi_j$ for each group, AxelGNN$_{\text{Sim}}$ maintains simple learnable parameters at the group level. Using the same grouping structure with $C = \lceil d^{(l)}/s \rceil$ groups, we maintain learnable parameters $\mathbf{W}_{\text{group}}^{(l)} \in \mathbb{R}^{C \times s}$ and compute group-specific copying probabilities:

\begin{equation}
\mathbf{C}^{(l)} = \sigma(\mathbf{W}_{\text{group}}^{(l)})
\end{equation}

where $\sigma: \mathbb{R}^{C \times s} \rightarrow (0,1)^{C \times s}$. For group $j$ containing traits $[js, \min((j+1)s, d^{(l)}))$, the copying probabilities are:

\begin{equation}
\mathbf{c}_j^{(l)} = \mathbf{C}^{(l)}[j, :(\text{group\_size})]
\end{equation}

The node update rule applies group-specific copying:
\begin{align}
\mathbf{x}_v^{(l)}[js\!:\!\min((j\!+\!1)s, d^{(l)})] 
&= \mathbf{c}_j^{(l)} \odot \mathbf{a}_v^{(l)}[js\!:\!\min((j\!+\!1)s, d^{(l)})] \notag \\
&\quad + (\mathbf{1} - \mathbf{c}_j^{(l)}) \odot \mathbf{h}_v^{(l)}[js\!:\!\min((j\!+\!1)s, d^{(l)})]
\end{align}

This simplification eliminates the need for group-specific neural networks $\phi_j$ while maintaining the same grouped copying structure. The group size hyperparameter $s$ continues to control the granularity of trait copying. The high-level architectural workflow in AxelGNN is depicted in Figure \ref{fig:axelgnn_architecture}.
\begin{figure}[htbp]
\centering
\begin{tikzpicture}[
  node/.style={rectangle, draw, minimum width=4cm, minimum height=1cm, align=center, font=\small},
  standard/.style={rectangle, draw, fill=blue!5, minimum width=4cm, minimum height=1cm, align=center, font=\small},
  novel/.style={rectangle, draw, fill=orange!20, minimum width=4cm, minimum height=1.2cm, align=center, font=\small, line width=1pt},
  arrow/.style={->, >=stealth, thick}
]


\node[node] (input) at (0,0) {Input: Node Features\\Adjacency Matrix};

\node[standard] (linear) at (0,-2.3) {Linear Transformation\\$h_u^{(l)} = W^{(l)}x_u^{(l-1)} + b^{(l)}$};

\node[novel] (similarity) at (0,-4.2) {
  \textbf{Similarity Calculation}\\
  $s_{vu}^{(l)} = \frac{h_v^{(l)} \cdot h_u^{(l)}}{\|h_v^{(l)}\|\|h_u^{(l)}\|}$
};

\node[novel] (interaction) at (0,-6.3) {
  \textbf{Interaction Probability}\\
  $p_{vu}^{(l)} = \sigma(\beta \cdot (s_{vu}^{(l)} - \theta))$\\
  {\footnotesize Bistable convergence}
};

\node[standard] (message) at (0,-8.4) {
  Message Passing\\
  $a_v^l = \frac{1}{|N(v)|}\sum_{u \in N(v)} p_{uv}^l \cdot h_u^l$
};

\node[novel] (segment) at (6.5,-9.35) {
  \textbf{Segment-wise Copying}\\
  Group-specific networks $\phi_j$\\
  $C = \lceil d^{(l)}/s \rceil$ groups
};

\node[standard] (updated) at (0,-10.3) {
  Updated Node Features\\
  $x_v^{(l)}$
};

\node[node] (output) at (0,-11.8) {
  Output: Node Embeddings\\
  (Global Polarization)
};

\draw[arrow] (input) -- (linear);
\draw[arrow] (linear) -- (similarity);
\draw[arrow] (similarity) -- (interaction);
\draw[arrow] (interaction) -- (message);
\draw[arrow] (message) -- (updated);
\draw[arrow] (updated) -- (output);

\draw[arrow] (message.east) -- (segment.west);
\draw[arrow] (segment.east) -- ++(0.6,0) |- (updated.east);

\draw[arrow, dashed, rounded corners]
  (updated.west) -- ++(-2.2,0) -- ++(0,8.0) -- (linear.west);
\node[anchor=east, font=\footnotesize, rotate=90] at (-4.5,-6.3) {Repeat for L layers};

\end{tikzpicture}
\caption{AxelGNN architecture overview showing the layer-wise processing flow with segment-based aggregation. Orange boxes highlight novel contributions: similarity-gated interaction, bistable convergence dynamics, and segment-wise trait copying.}
\label{fig:axelgnn_architecture}
\end{figure}

\subsection{Comparison of AxelGNN with Axelrod Cultural Dissemination Model}

In this section, we compare different components of the Axelrod cultural dissemination model with our AxelGNN architecture. This comparison demonstrates that our GNN offers a robust neural approximation of the Axelrod model by retaining its key components, thereby extending its stochastic and discrete nature into a differentiable and continuous setting, which is required for learning tasks.

\begin{table}[h!]
\centering
\begin{tabular}{l|p{5cm}|p{5cm}}
\hline
Component & Axelrod Model & AxelGNN \\
\hline
Mechanism & Discrete copying & Neural gating \\
\hline
Learning & Static rules & Trainable parameters \\
\hline
Traits & Discrete features & Continuous features \\
\hline
Similarity & Trait overlap & Feature similarity \\
\hline
Interaction Rule & Probabilistic & Learned function \\
\hline
Trait Adoption & Copy one random trait from neighbor & Weighted aggregation + feature copying \\
\hline
Scope & Local & Local  \\
\hline
Dynamics & Stochastic and agent-based & Deterministic and differentiable\\
\hline
\end{tabular}
\caption{Axelrod Model vs AxelGNN Comparison}
\label{tab:axelrod_comparison}
\end{table}

\subsection{Distinction from Attention and Gating Mechanisms}

AxelGNN fundamentally differs from attention and gating mechanisms \citep{velivckovic2018graph,ruiz2020gated}  in multiple ways. Its bistable convergence dynamics, which employ sharp thresholding, differ from those of continuous weights in attention and gating models. The global polarisation constraints in AxelGNN maintain distinct clusters, unlike independent local weight mechanisms. Further, our segment-wise trait copying employs semantic group operations that are fundamentally different from those of existing feature-wise gating models. These design choices, grounded in Axelrod's model, provide principled oversmoothing prevention absent in standard attention and gating mechanisms.
Table~\ref{tab:mechanism_comparison} summarizes the key architectural differences between AxelGNN and existing attention and gating approaches.

\begin{table}[h]
\centering
\small
\begin{tabular}{lcccc}
\toprule
Mechanism & Weight & Decision & Scope & Oversmoothing \\
 & Type & Boundary &  & Prevention \\
\midrule
Attention & Continuous & Smooth sigmoid & Local Neighborhood & None \\
Gating & Continuous & Smooth sigmoid & Features & Via residual \\
AxelGNN & Bistable & Sharp threshold & Global Structure & Via polarization \\
\bottomrule
\end{tabular}
\caption{Comparison of AxelGNN with graph attention and gating mechanisms}
\label{tab:mechanism_comparison}
\end{table}

\subsection{Complexity Analysis} 

The computational complexity varies between the two variants of our model. In AxelGNN, the computational complexity at layer $l$ is:
\[
\mathcal{O}(|V|d^{(l-1)}d^{(l)} + |E|d^{(l)} + |V| \cdot C \cdot s \cdot k)
\]
where $C = \lceil d^{(l)}/s \rceil$ is the number of groups, $s$ is the group size, and $k$ represents the hidden dimension of the group-specific neural networks $\phi_j$. The term $|V| \cdot C \cdot s \cdot k$ is associated with calculating group-specific copying probabilities. 

AxelGNN$_{\text{Sim}}$ simplifies the complexity of AxelGNN to:
\[
\mathcal{O}(|V|d^{(l-1)}d^{(l)} + |E|d^{(l)})
\]
which removes the expensive group-specific neural network computations. The parameter count for copying mechanisms is reduced from $C \times k$ neural network parameters per group to $C \times s$ simple weights, achieving a significant speedup factor of approximately $|V|k$ in the computation of copying probabilities. The choice of group size $s$ becomes a key hyperparameter that allows us to tune the trade-off between model expressiveness and computational efficiency. This simplification makes AxelGNN$_{\text{Sim}}$ particularly well-suited for large-scale graphs where computational efficiency is crucial.

\section{Experimental Design}

To evaluate AxelGNN's ability to model complex relationships, we employ two downstream tasks: node classification and influence estimation. We acknowledge that these tasks do not represent the full spectrum of graph learning challenges. To comprehensively validate the generalizability of our approach, we plan to evaluate AxelGNN on structurally different tasks, including link prediction, graph classification, and graph regression tasks, in our future work.

\subsection{Downstream tasks}

\subsubsection{Node Classification}  Node classification aims to predict a class label for each node in the graph \cite{xiao2022graph}. GNN modeling of complex relationships is essential for node classification, as some real-world datasets exhibit heterophily, making traditional aggregation ineffective. Moreover, oversmoothing results in loss of node discriminability, making them indistinguishable.

\subsubsection{Influence Estimation for Epidemic Modeling} Influence estimation predicts activation probabilities for nodes given initial seeds and a diffusion model, with applications to epidemic forecasting, viral marketing, and information spread \cite{xia2021deepis}. Traditional algorithms rely on computationally expensive Monte Carlo simulations, making them impractical for large-scale networks or time-critical applications \cite{xia2021deepis}. GNN-based methods reduce complexity from exponential to linear by learning probabilities directly from network structure. We evaluate on both progressive (Linear Threshold) and non-progressive (Susceptible-Infected-Susceptible) diffusion models, capturing scenarios from one-time adoption to cyclical epidemic dynamics. Modeling complex relationships in influence estimation is critical because diffusion involves multi-hop dependencies where spreading occurs both within similar groups (homophily) and across boundaries (heterophily). Avoiding oversmoothing is essential since diffusion modeling requires deeper GNNs ( more than 2-4 layers) to capture multi-step propagation dynamics through multiple degrees of separation.

\subsection{Datasets}

For node classification, we employ eight datasets consisting of both homophilic and heterophilic datasets. Homophilic datasets include Cora, Citeseer, and Pubmed from CitationFull benchmark \cite{yang2016revisiting}, and heterophilic datasets include Cornell, Film, Wisconsin, and Texas from WebKB \cite{pei2020geom}, as well as Penn94 from LINKX \cite{lim2021large}. For influence estimation, we employ four real-world datasets, namely, Jazz, Cora-ML, Network Science, and Power Grid \cite{rossi2015network,bojchevski2018deep}. The dataset statistics for node classification and influence estimation are provided in Table \ref{tab:node_classification_datsets} and Table \ref{tab:influence_estimation_datasets}, respectively. Additionally, we employ ogbn-arxiv dataset from OGB benchmark \cite{hu2020open} for the scalability analysis.

\begin{table}[h]
\centering
\scalebox{0.9}{
\begin{tabular}{l|r|r|r|r|r}
\hline
Datasets & Nodes & Edges & Features & Classes & Homophily Ratio \\
\hline
Cora & 2708 & 5278 & 1433 & 7 & 0.81 \\
Pubmed & 19,717 & 44,327 & 500 & 3 & 0.80 \\
Citeseer & 3327 & 4676 & 3703 & 6 & 0.74 \\
Penn94 & 41,554 & 1,362,229 & 5 & 2 & 0.47 \\
Cornell & 183 & 280 & 1703 & 5 & 0.30 \\
Film & 7600 & 26,752 & 931 & 5 & 0.22 \\
Wisconsin & 251 & 466 & 1,703 & 5 & 0.21 \\
Texas & 183 & 295 & 1703 & 5 & 0.11 \\
\hline
ogbn-arxiv & 169,343 & 1,166,243 & 128 & 40 & 0.65 \\
\hline
\end{tabular}
}
\caption{Dataset Statistics - Node Classification}
\label{tab:node_classification_datsets}
\end{table}

\begin{table}[h]
\centering
\scalebox{0.9}{
\begin{tabular}{l|r|r}
\hline
Dataset & Nodes & Edges \\
\hline
Jazz & 198 & 2,742 \\
Cora-ML & 2,810 & 7,981 \\
Network Science & 1,565 & 13,532 \\
Power Grid & 4,941 & 6,594 \\
\hline
\end{tabular}
}
\caption{Dataset Statistics - Influence Estimation}
\label{tab:influence_estimation_datasets}
\end{table}

\subsection{Baselines}

For node classification, we compare against baselines organized into four categories:
\begin{itemize}
    \item \textbf{Standard GNNs:} GCN~\cite{kipf2017semi}, GAT~\cite{velivckovic2018graph}, and GraphSAGE~\cite{hamilton2017inductive}.
    \item \textbf{Oversmoothing-aware methods:} MixHop~\cite{abu2019mixhop} and GCNII~\cite{chen2020simple}.
    \item \textbf{Heterophily-oriented methods:} H2GCN~\cite{zhu2020beyond}, WRGAT~\cite{suresh2021breaking}, ACM-GCN~\cite{luan2022revisiting}, LINKX~\cite{lim2021large}, GloGNN++~\cite{li2022finding}, PCNet~\cite{li2024pc}, and GNRF~\cite{chen2025graph}.
    \item \textbf{Hybrid approaches:} GPRGNN~\cite{chienadaptive}, GGCN~\cite{yan2022two}, DirGNN~\cite{rossi2024edge}, and BEC-GCN~\cite{hevapathige2025depth}.
\end{itemize}

For influence estimation, we compare against baselines organized into two categories:
\begin{itemize}
    \item \textbf{Standard GNNs:} GCN~\cite{kipf2017semi}, GAT~\cite{velivckovic2018graph}, and GraphSAGE~\cite{hamilton2017inductive}.
    \item \textbf{Influence estimation methods:} DeepIS~\cite{xia2021deepis}, DeepIM~\cite{ling2023deep}, GLIE~\cite{panagopoulos2023maximizing}, and UniGO~\cite{li2025unigo}.
\end{itemize}

\subsection{Evaluation Settings}

For node classification, we employ a training/validation/testing split of 60/20/20 following work in literature \cite{suresh2021breaking,li2024pc}. For AxelGNN variants and baselines, we report the mean and standard deviation of accuracy over 10 random initialisations. 

We evaluate influence estimation over two diffusion models: Linear Threshold (LT) \cite{kempe2003maximizing}, and Susceptible-Infected-Susceptible (SIS) \cite{kimura2009efficient}. LT is a progressive diffusion model where nodes become activated when the weighted influence from their activated neighbors exceeds a predefined threshold, and SIS is a non-progressive diffusion model where nodes can transition between susceptible and infected states, allowing previously infected nodes to become susceptible again. We follow the experimental setup and dataset splits provided by Ling et al. \cite{ling2023deep}. In our experiments, we randomly select 10\% of nodes as the initial activation set for each diffusion scenario, use 10-fold cross-validation and report both the mean and standard deviation of Mean Absolute Error (MAE).

For baseline results, we employ results reported in previous papers that have the same experimental setups. When such results are not available, we run the baseline methods using the hyperparameter settings specified in their original papers and report the results to ensure fair comparison. 

\subsection{Model Hyperparameters}

We employ grid search \cite{liashchynskyi2019grid} to find the best hyperparameters for AxelGNN variants. We use different numbers of layers $\in \{1, 2, 3, 4\}$, learning rates $\in \{1e^{-3}, 5e^{-3}, 1e^{-2}\}$, weight decay values $\in \{5e^{-4}, 1e^{-1}\}$, dropout rates $\in \{0.1, 0.4, 0.5\}$, hidden dimensions $\in \{32, 64, 256\}$, and segment sizes (s) $\in \{4, 8, 16\}$. Models were trained for up to 1000 epochs for node classification and 200 epochs for influence estimation. Further, we employ the Adam algorithm \cite{kingma2014adam} for model optimization.

\subsection{System Resources, and Implementation Details}

We conduct all our experiments on a Linux server equipped with an Intel Xeon W-2175 2.50 GHz processor across 28 cores, with an NVIDIA RTX A6000 GPU and 512 GB of RAM. For implementation, we use the Python programming language \cite{sanner1999python} with PyTorch 2.3.1 as the core framework, along with the following libraries: torchvision 0.18.1, torchaudio 2.3.1, torch-geometric 2.7.0, torch-cluster 1.6.3, torch-scatter 2.0.9, and torch-sparse 0.6.18 \cite{paszke2019pytorch}.

\section{Results, and Discussion}

\subsection{How do AxelGNN variants compare to existing GNNs for node classification and influence estimation benchmarks?}

To evaluate the effectiveness of our proposed AxelGNN variants, we conduct comprehensive experiments comparing their performance against existing GNN methods across node classification and influence estimation benchmarks.

\subsubsection{Node Classification}

\begin{table}[h!]
\centering
\scalebox{0.725}{
\begin{tabular}{l|c|c|c|c|c|c|c|c}
\hline
\textbf{Dataset} & \textbf{Cora} & \textbf{Citeseer} & \textbf{Pubmed} & \textbf{Texas} & \textbf{Actor} & \textbf{Cornell} & \textbf{Wisconsin} & \textbf{Penn94} \\
\hline
GCN & 86.98$_{\pm 1.27}$ & 76.50$_{\pm 1.36}$ & 88.42$_{\pm 0.50}$ & 55.14$_{\pm 5.16}$ & 27.32$_{\pm 1.10}$ & 60.54$_{\pm 5.30}$ & 51.76$_{\pm 3.06}$ & 82.47$_{\pm 0.27}$ \\
GAT & 87.30$_{\pm 1.10}$ & 76.55$_{\pm 1.23}$ & 86.33$_{\pm 0.48}$ & 52.16$_{\pm 6.63}$ & 27.44$_{\pm 0.89}$ & 61.89$_{\pm 5.05}$ & 49.41$_{\pm 4.09}$ & 81.59$_{\pm 0.55}$ \\
GraphSAGE & 86.90$_{\pm 1.04}$ & 76.04$_{\pm 1.30}$ & 88.45$_{\pm 0.50}$ & 82.43$_{\pm 6.14}$ & 34.23$_{\pm 0.99}$ & 75.95$_{\pm 5.01}$ & 81.18$_{\pm 5.56}$ & 80.94$_{\pm 0.66}$ \\
\hline
MixHop & 87.61$_{\pm 0.85}$ & 76.26$_{\pm 1.33}$ & 85.31$_{\pm 0.61}$ & 77.84$_{\pm 7.73}$ & 32.22$_{\pm 2.34}$ & 73.51$_{\pm 6.34}$ & 75.88$_{\pm 4.90}$ & 83.40$_{\pm 0.71}$ \\
GCNII & 88.37$_{\pm 1.25}$ & 77.33$_{\pm 1.48}$ & 90.15$_{\pm 0.43}$ & 77.57$_{\pm 1.83}$ & 37.44$_{\pm 1.30}$ & 77.86$_{\pm 3.79}$ & 80.39$_{\pm 3.40}$ & 82.92$_{\pm 0.59}$ \\
H2GCN & 87.87$_{\pm 1.20}$ & 77.11$_{\pm 1.57}$ & 89.49$_{\pm 0.38}$ & 84.86$_{\pm 2.73}$ & 35.70$_{\pm 4.00}$ & 82.70$_{\pm 5.28}$ & 87.65$_{\pm 4.98}$ & 81.31$_{\pm 0.60}$ \\
WRGAT & 88.20$_{\pm 1.26}$ & 76.81$_{\pm 1.89}$ & 88.52$_{\pm 0.99}$ & 83.62$_{\pm 6.50}$ & 36.33$_{\pm 1.77}$ & 81.62$_{\pm 3.90}$ & 86.98$_{\pm 3.78}$ & 74.32$_{\pm 1.63}$ \\
GPRGNN & 87.95$_{\pm 1.18}$ & 77.13$_{\pm 1.67}$ & 87.54$_{\pm 0.38}$ & 78.38$_{\pm 4.36}$ & 34.63$_{\pm 1.22}$ & 80.27$_{\pm 8.11}$ & 82.94$_{\pm 4.21}$ & 81.38$_{\pm 0.16}$ \\
GGCN & 87.95$_{\pm 1.05}$ & 77.14$_{\pm 1.45}$ & 89.15$_{\pm 0.37}$ & 84.86$_{\pm 4.55}$ & 37.54$_{\pm 1.56}$ & 85.68$_{\pm 6.63}$ & 86.86$_{\pm 3.22}$ & OOM \\
ACM-GCN & 87.91$_{\pm 0.95}$ & 77.32$_{\pm 1.70}$ & 90.00$_{\pm 0.52}$ & 87.84$_{\pm 4.40}$ & 36.28$_{\pm 1.09}$ & 85.14$_{\pm 6.07}$ & 88.43$_{\pm 3.22}$ & 82.52$_{\pm 0.96}$ \\
LINKX & 84.64$_{\pm 1.13}$ & 73.19$_{\pm 0.99}$ & 87.86$_{\pm 0.77}$ & 74.60$_{\pm 8.37}$ & 36.10$_{\pm 1.55}$ & 77.84$_{\pm 5.81}$ & 75.49$_{\pm 5.72}$ & 84.71$_{\pm 0.52}$ \\
GloGNN++ & 88.33$_{\pm 1.09}$ & 77.22$_{\pm 1.78}$ & 89.24$_{\pm 0.39}$ & 84.05$_{\pm 4.90}$ & 37.70$_{\pm 1.40}$ & 85.95$_{\pm 5.10}$ & 88.04$_{\pm 3.22}$ & 85.74$_{\pm 0.42}$ \\
PCNet & \textbf{88.41$_{\pm 0.66}$} & 77.50$_{\pm 1.06}$ & 89.51$_{\pm 0.28}$ & 88.11$_{\pm 2.17}$ & 37.80$_{\pm 0.64}$ & 82.16$_{\pm 2.70}$ & 88.63$_{\pm 2.75}$ & 84.75$_{\pm 0.51}$ \\
DirGNN & 86.78$_{\pm 0.97}$ & 77.71$_{\pm 0.78}$ & 86.94$_{\pm 0.55}$ & 76.25$_{\pm 4.68}$ & 35.76$_{\pm 6.31}$ & 76.51$_{\pm 6.14}$ & 80.50$_{\pm 5.50}$ & 79.58$_{\pm 0.75}$ \\
GNRF & 87.99$_{\pm 1.02}$ & 75.79$_{\pm 0.94}$ & \textbf{90.37$_{\pm 0.69}$} &  87.39$_{\pm 4.13}$ & 34.22$_{\pm 1.40}$ & \textbf{87.28$_{\pm 3.12}$} & 88.00$_{\pm 2.00}$ & 84.88$_{\pm 0.85}$ \\
BEC-GCN & 88.21$_{\pm 0.80}$ & 78.14$_{\pm 1.31}$ & 87.00$_{\pm 0.28}$ &  68.85$_{\pm 11.02}$ & 34.34$_{\pm 2.08}$ & 65.96$_{\pm 9.03}$ & 66.25$_{\pm 4.55}$ & 85.97$_{\pm 0.95}$ \\
\hline
AxelGNN & 88.18$_{\pm 1.12}$ & 77.98$_{\pm 1.12}$ & 89.83$_{\pm 0.66}$ & \textbf{88.36$_{\pm 5.00}$} & \textbf{39.30$_{\pm 1.18}$} & 87.02$_{\pm 3.49}$ & \textbf{91.25$_{\pm 2.02}$} & \textbf{86.01$_{\pm 1.00}$} \\
AxelGNN$_\text{Sim}$ & 87.03$_{\pm 1.04}$ & \textbf{78.25$_{\pm 1.27}$} & 88.92$_{\pm 0.46}$ & 86.23$_{\pm 3.75}$ & 37.64$_{\pm 1.69}$ & 84.04$_{\pm 4.97}$ & 87.63$_{\pm 3.60}$ & 85.98$_{\pm 0.97}$ \\
\hline
\end{tabular}
}
\caption{Node classification accuracy ± standard deviation (\%). The best results are highlighted in \textbf{bold}. OOM refers to the out-of-memory. }
\label{tab:node_classification_comparison}
\end{table}

AxelGNN model performance comparison with base GNNs for the node classification task is shown in Table \ref{tab:node_classification_comparison}. As per results, AxelGNN consistently outperforms or performs competitively with the performance of both traditional and heterophily-aware GNNs. We attribute this to AxelGNN's ability to model bistable convergence dynamics with heterogeneous relationships between neighbouring nodes, enabling the successful modelling of complex structural connections within both homophilic and heterophilic label patterns. Additionally, AxelGNN$_\text{Sim}$ also provides comparably solid performance, demonstrating its ability to become an effective and computationally efficient solution for real-world datasets.

\subsubsection{Influence Estimation}

AxelGNN performance comparison for the influence estimation task is depicted in Table \ref{tab:influence_estimation_comparison}. Both AxelGNN variants consistently outperform traditional GNNs, as well as GNNs designed for the influence estimation task. Influence estimation requires the GNN to adapt to the temporal dynamics of the diffusion model. The similarity-gated behaviour of AxelGNN is dynamically changing with each time step, helping AxelGNN variants to mimic the underlying diffusion patterns successfully. Similar to node classification,  AxelGNN$_\text{Sim}$ also provides a solid performance for the influence estimation task, demonstrating its strength as an effective and scalable solution.

\begin{table}[h!]
\centering
\scalebox{0.7}{
\begin{tabular}{l|cc|cc|cc|cc}
\hline
& \multicolumn{2}{c|}{\textbf{Jazz}} & \multicolumn{2}{c|}{\textbf{Cora-ML}} & \multicolumn{2}{c|}{\textbf{Network Science}} & \multicolumn{2}{c}{\textbf{Power Grid}} \\
& \textbf{LT} & \textbf{SIS} & \textbf{LT} & \textbf{SIS} & \textbf{LT} & \textbf{SIS} & \textbf{LT} & \textbf{SIS} \\
\hline
GCN & 0.199$_{\pm 0.006}$ & 0.344$_{\pm 0.023}$ & 0.255$_{\pm 0.008}$ & 0.365$_{\pm 0.065}$ & 0.190$_{\pm 0.012}$ & 0.180$_{\pm 0.007}$ & 0.335$_{\pm 0.023}$ & 0.207$_{\pm 0.001}$ \\
GAT & 0.156$_{\pm 0.100}$ & \textbf{0.288$_{\pm 0.017}$} & 0.192$_{\pm 0.010}$ & 0.208$_{\pm 0.008}$ & 0.114$_{\pm 0.008}$ & 0.123$_{\pm 0.013}$ & 0.280$_{\pm 0.015}$ & 0.133$_{\pm 0.001}$ \\
GraphSAGE & 0.120$_{\pm 0.004}$ & 0.301$_{\pm 0.018}$ & 0.203$_{\pm 0.009}$ & 0.222$_{\pm 0.051}$ & 0.112$_{\pm 0.005}$ & 0.102$_{\pm 0.010}$ & 0.341$_{\pm 0.018}$ & 0.133$_{\pm 0.001}$\\ 
\hline
DeepIS & 0.219$_{\pm 0.002}$ & 0.434$_{\pm 0.003}$ & 0.301$_{\pm 0.005}$ & 0.304$_{\pm 0.001}$ & 0.306$_{\pm 0.001}$ & 0.256$_{\pm 0.001}$ & 0.374$_{\pm 0.001}$ & 0.251$_{\pm 0.001}$ \\
DeepIM & 0.134$_{\pm 0.014}$ & 0.383$_{\pm 0.010}$ & 0.271$_{\pm 0.010}$ & 0.270$_{\pm 0.006}$ & 0.118$_{\pm 0.003}$ & 0.135$_{\pm 0.004}$ & 0.331$_{\pm 0.002}$ & 0.205$_{\pm 0.005}$ \\
GLIE & 0.055$_{\pm 0.028}$ & 0.454$_{\pm 0.062}$ & 0.286$_{\pm 0.016}$ & 0.205$_{\pm 0.029}$ & 0.160$_{\pm 0.003}$ & 0.103$_{\pm 0.023}$ & 0.384$_{\pm 0.020}$ & \textbf{0.132$_{\pm 0.026}$} \\
UniGO & 0.159$_{\pm 0.020}$ & 0.335$_{\pm 0.002}$ & 0.155$_{\pm 0.019}$ & 0.212$_{\pm 0.001}$ & 0.110$_{\pm 0.049}$ & 0.108$_{\pm 0.023}$ & 0.235$_{\pm 0.005}$ & 0.206$_{\pm 0.023}$ \\
\hline
AxelGNN & \textbf{0.051$_{\pm 0.002}$} & 0.321$_{\pm 0.006}$ & \textbf{0.144$_{\pm 0.008}$} & 0.197$_{\pm 0.003}$ & \textbf{0.046$_{\pm 0.002}$} & \textbf{0.100$_{\pm 0.002}$} & \textbf{0.223$_{\pm 0.003}$} & 0.135$_{\pm 0.004}$ \\
AxelGNN$_\text{Sim}$ & 0.054$_{\pm 0.002}$ & 0.320$_{\pm 0.006}$ & 0.152$_{\pm 0.005}$ & \textbf{0.195$_{\pm 0.001}$} & 0.060$_{\pm 0.003}$ & 0.103$_{\pm 0.002}$ & 0.229$_{\pm 0.002}$ & 0.137$_{\pm 0.001}$ \\
\hline
\end{tabular}
}
\caption{Influence estimation MAE ± standard deviation (\%). Lower value indicates better performance. The best results are highlighted in \textbf{bold}.}
\label{tab:influence_estimation_comparison}
\end{table}

 \subsection{How robust are AxelGNN variants to oversmoothing across different network depths?}

 \begin{figure}[htbp]
    \centering
    \includegraphics[width=0.9\textwidth]{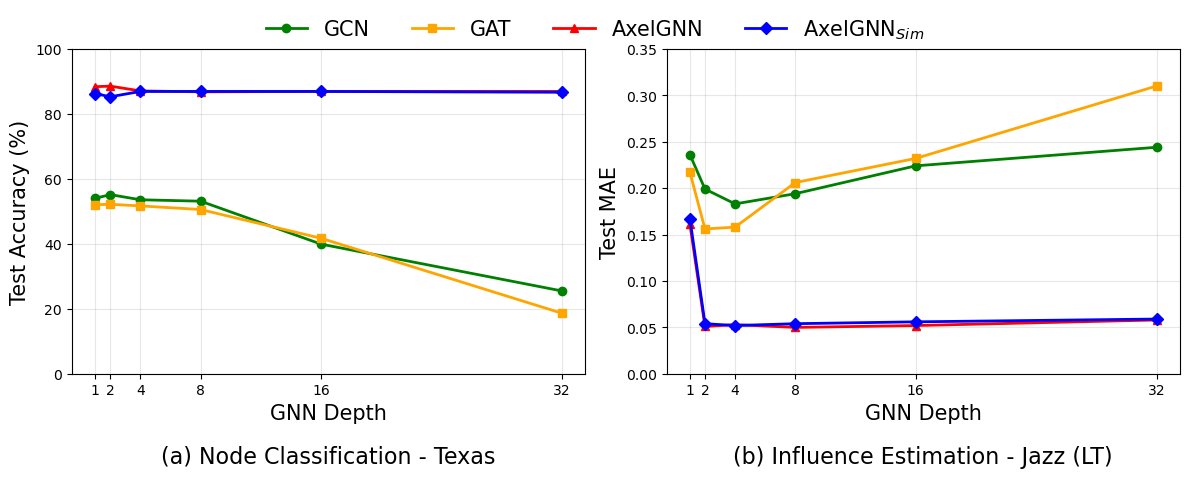}
    \caption{Robustness comparison of AxelGNN variants against oversmoothing across different network depths.}
    \label{fig:oversmoothing}
\end{figure}

Figure \ref{fig:oversmoothing} demonstrates the robustness of AxelGNN variants for oversmoothing. For node classification tasks, AxelGNN variants exhibit stable performance across different network depths, while traditional GNNs such as GCN and GAT show significant performance degradation as the number of layers increases. In influence estimation benchmarks, AxelGNN variants consistently maintain low estimation error across varying depths, while conventional GNN architectures experience a substantial increase in estimation error in deeper networks. We attribute this to the global polarization mechanism in our approach that maintains embedding diversity across nodes, effectively preventing the homogenization of node representations.

\subsection{What is the impact of hyperparameters on AxelGNN's performance-efficiency trade-offs?}

\begin{figure}[htbp]
    \centering
    \includegraphics[width=\textwidth]{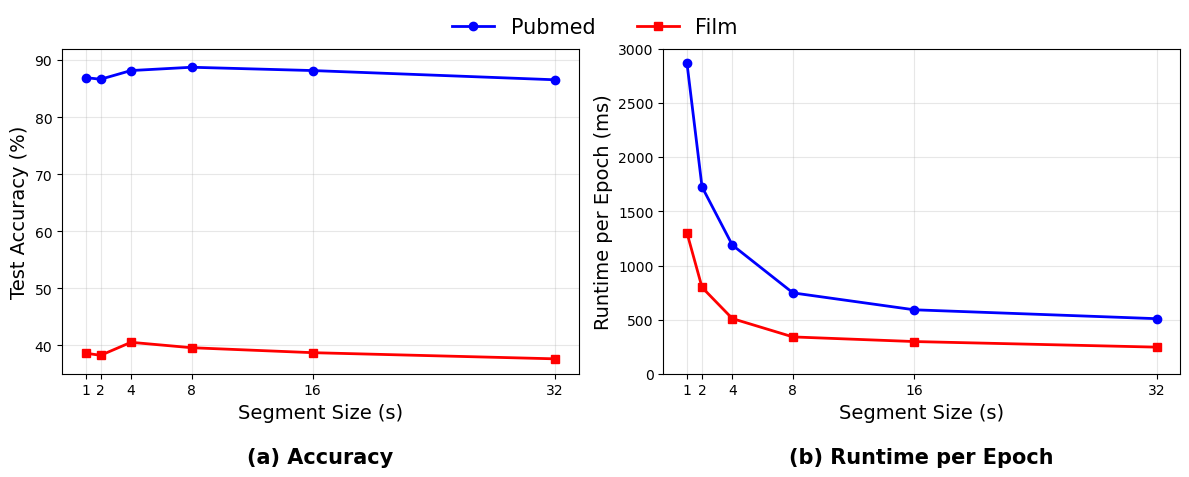}
    \caption{Parameter sensitivity analysis of segment size (s) in AxelGNN across Pubmed and Film datasets.}
    \label{fig:segment_size_sensitivity}
\end{figure}

We evaluate the performance-efficiency trade-offs of AxelGNN concerning its key hyperparameter, segment size (s), as shown in Figure \ref{fig:segment_size_sensitivity}. The results indicate that optimal performance is achieved when the segment size (s) is set to moderate values between 4 and 8. This range strikes an effective balance between model expressiveness and computational efficiency.

 Using very small values, such as s = 1 or 2, causes the model to become overly granular, treating each feature individually and disrupting important relationships between them that are crucial for learning. This leads to poorer performance. Conversely, when s is too large, like s = 32, the copying mechanism becomes excessively broad and combines different types of features, making it difficult for the model to make precise decisions about which features to copy. This also negatively impacts performance. Overall, this analysis shows that moderate segment sizes provide the best trade-off. They maintain enough granularity for effective feature copying while avoiding the computational burden of excessive fine-grained operations and the loss of expressiveness from overly coarse groupings.

The threshold parameter $\theta$ and intensity parameter $\beta$ are learned automatically during training instead of being manually tuned. These parameters are optimized through backpropagation, adapting to ground truth labels via the downstream loss function. This allows the model to discover optimal interaction dynamics for specific datasets and tasks, with $\theta$ determining the similarity threshold for node interactions and $\beta$ defining the steepness of the interaction probability function.

\subsection{How do AxelGNN variants balance computational efficiency with classification performance?}

\begin{figure}[H]
    \centering
    \includegraphics[width=\textwidth]{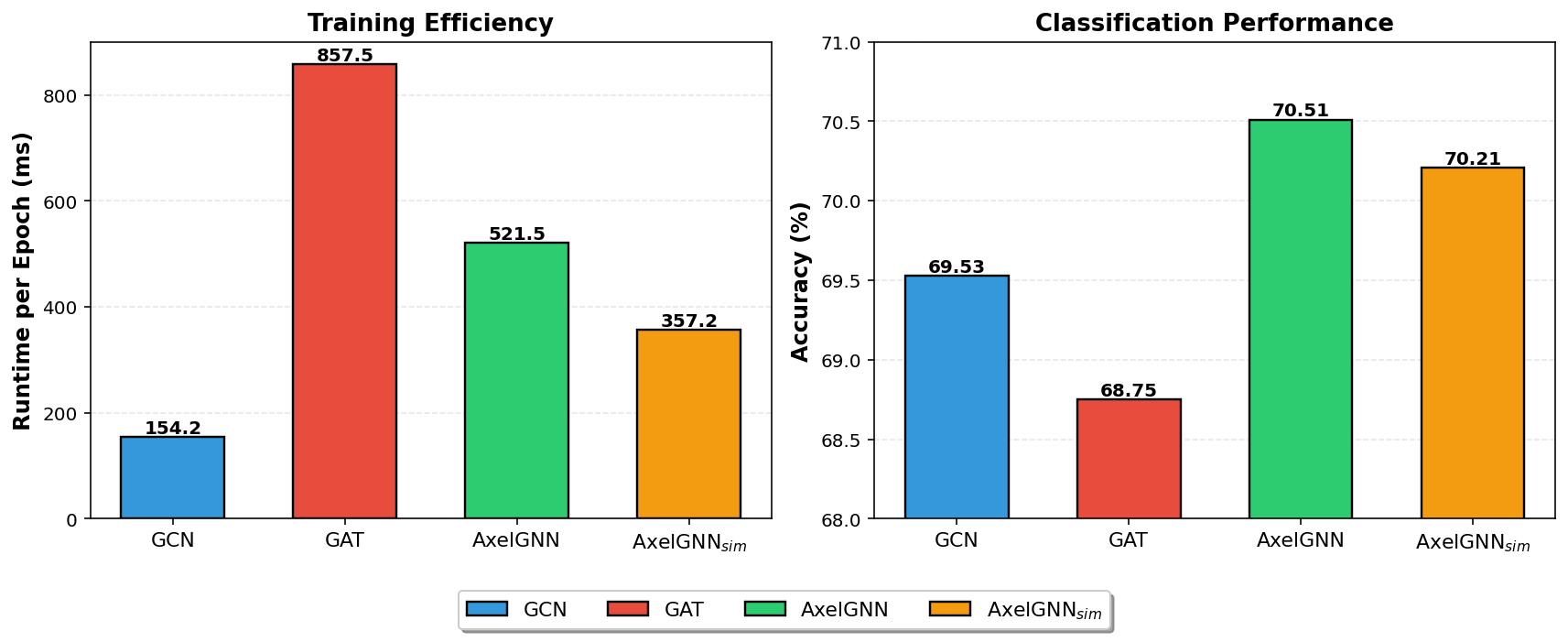}
    \caption{Runtime efficiency and classification performance comparison on ogbn-arxiv dataset for node classification.}
    \label{fig:scalability}
\end{figure}

To ensure practical applicability, we evaluate the computational efficiency of AxelGNN variants compared to baseline methods. For fair comparison, all models are employed with the same capacity (i.e., number of hidden layers and feature dimensions). Figure~\ref{fig:scalability} presents runtime efficiency and classification accuracy on large-scale ogbn-arxiv dataset for node classification task. AxelGNN achieves superior accuracy while requiring substantially less training time than GAT. AxelGNN$_\text{Sim}$ demonstrates even stronger efficiency gains while maintaining competitive accuracy. These results confirm that AxelGNN variants achieve better accuracy-efficiency tradeoffs. The efficiency advantage stems from our single global similarity computation combined with segment-wise copying, which proves more efficient than multi-head attention mechanisms. This positions AxelGNN as a practical solution for large-scale graph learning tasks.

\subsection{How do AxelGNN variants exhibit convergence dynamics? }

\begin{figure}[htbp]
    \centering
    \includegraphics[width=\textwidth]{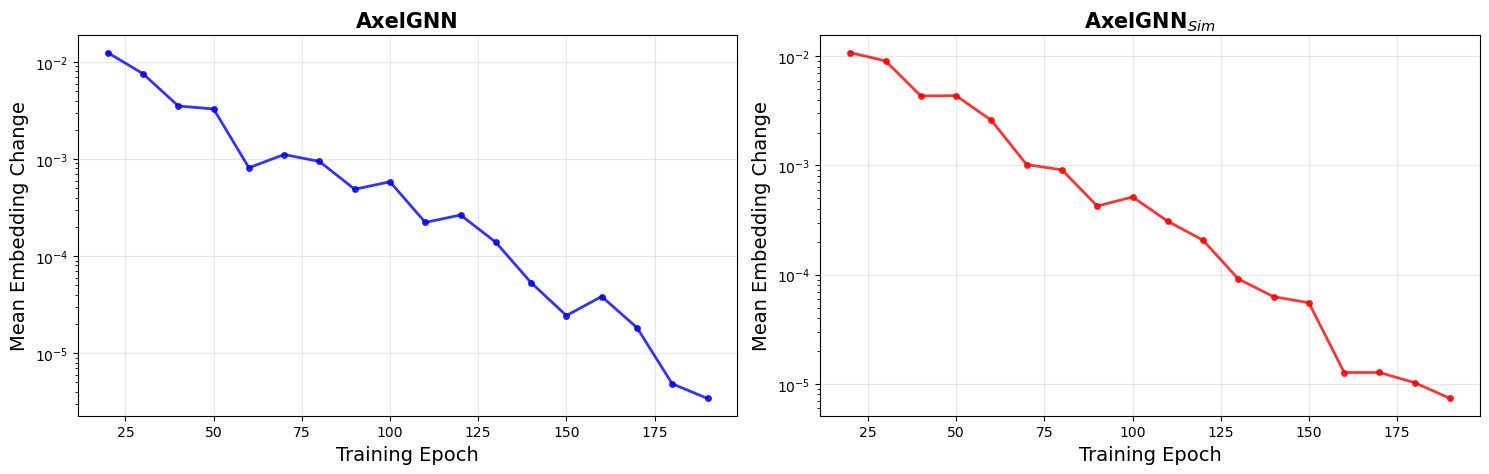}
    \caption{Embedding convergence dynamics of AxelGNN variants. }
    \label{fig:variant_convergence}
\end{figure}

To demonstrate the embedding convergence dynamics of the AxelGNN variants, we plot the mean embedding changes throughout the model learning process on the Citeseer node classification benchmark. The plots are shown in  Figure \ref{fig:variant_convergence}.

The exponential decay and subsequent plateau of the curves signify that both models are successfully learning stable representations and achieving embedding convergence. This demonstrates AxelGNN adheres to the theoretical property of convergence in Axelrod's cultural dissemination model. Further, AxelGNN achieves a lower and smoother convergence, indicating a more stable and efficient optimization trajectory compared to the AxelGNN$\mathbf{_{sim}}$ variant. This improvement can be attributed to a more detailed and precise parameter design in AxelGNN.

\begin{figure}[htbp]
    \centering
    \includegraphics[width=\textwidth]{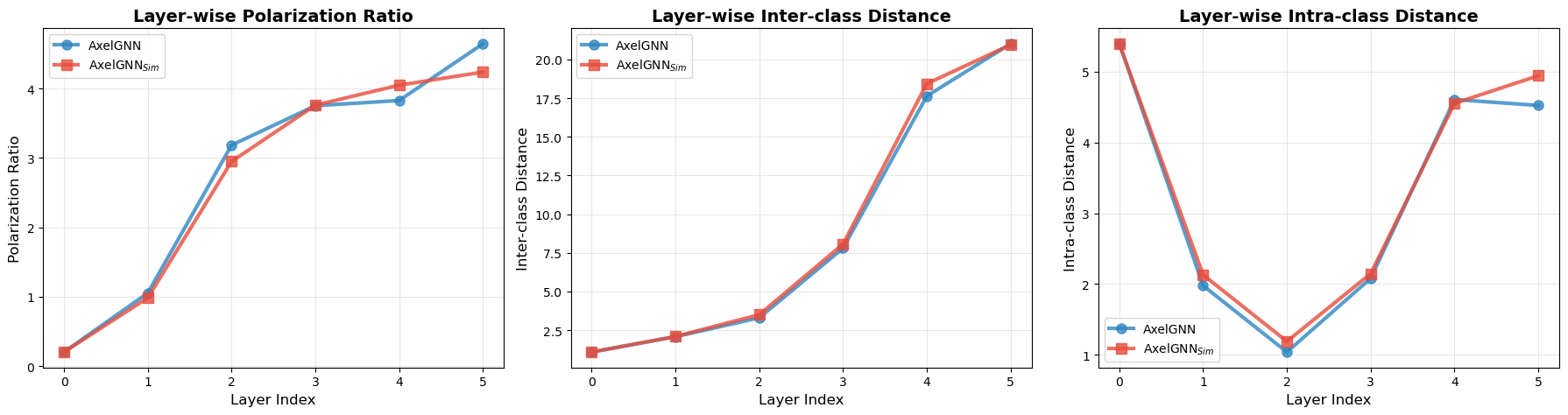}
    \caption{Layer-wise polarization dynamics across network depth for AxelGNN variants.}
    \label{fig:class_polarization}
\end{figure}

\subsection{How do AxelGNN variants exhibit global polarization dynamics across network layers?}

To evaluate the global polarization property of AxelGNN variants, we analyze their layer-wise polarization dynamics for node classification on the Citeseer dataset. Figure \ref{fig:class_polarization} depicts these dynamics, showing (a) polarization ratio progression (inter-class distance / intra-class distance), (b) inter-class distance evolution, and (c) intra-class distance changes through layers. Both AxelGNN variants exhibit polarization development patterns with the polarization ratio increasing monotonically, indicating progressive class separation. The inter-class distance grows rapidly, indicating node embeddings in different classes would have proper distance separation. 

\begin{figure}[htbp]
    \centering
    \includegraphics[width=\textwidth]{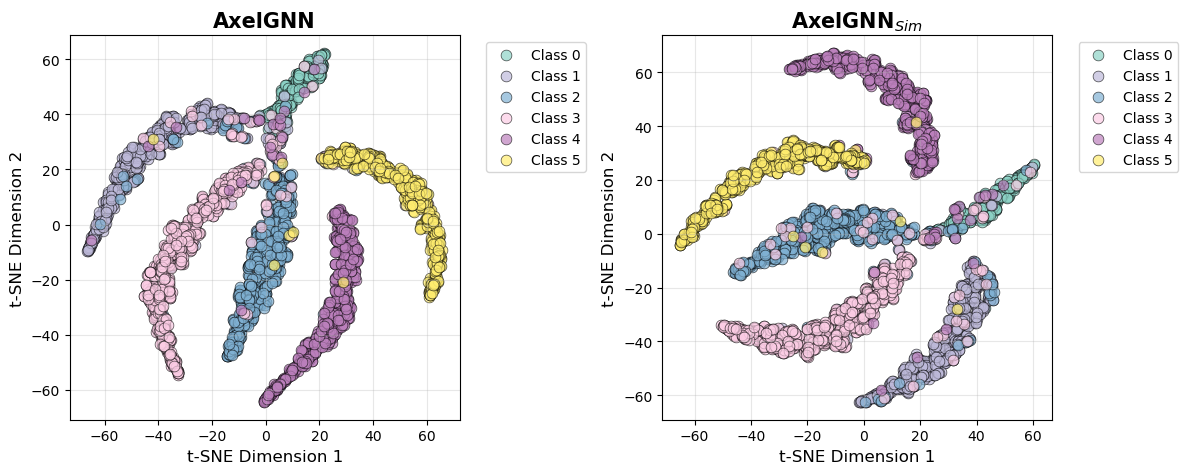}
    \caption{Final embedding space visualization using t-SNE projection for AxelGNN variants.}
    \label{fig:embedding_polarization}
\end{figure}

The t-SNE visualisations of embeddings in Figure \ref{fig:embedding_polarization} reveal successful class separation in both variants. Both achieve comparable cluster separation without inter-class overlap, suggesting evidence on the ability of AxelGNN variants to achieve global polarization without suffering from oversmoothing. 

\subsection{How robust are AxelGNN variants in maintaining distinct clustering patterns across different network depths?}

\begin{figure}[htbp]
    \centering
    
    \subfloat[AxelGNN]{%
        \includegraphics[width=\textwidth]{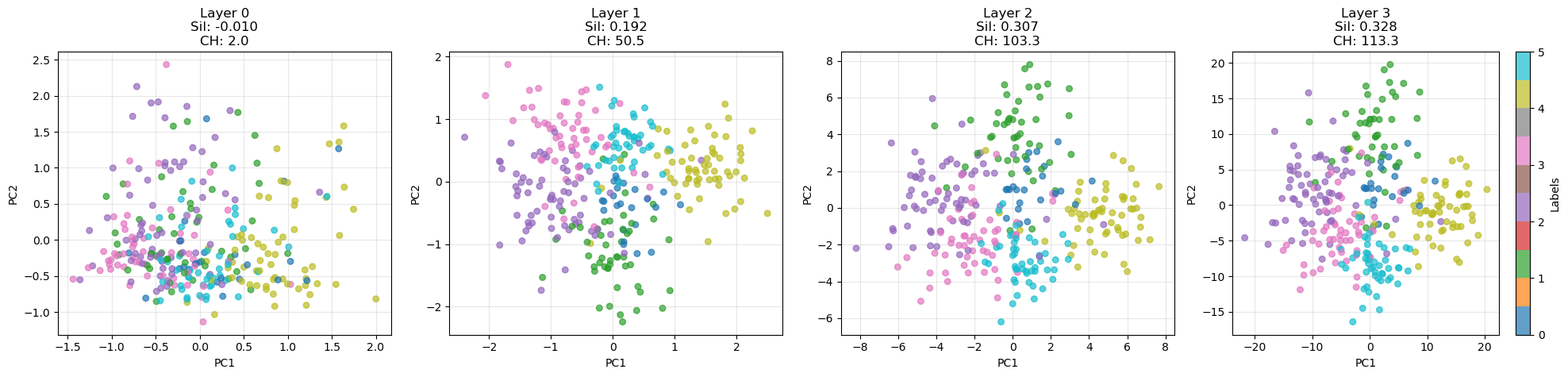}
        \label{fig:axelgnn_clustering}
    }
    
    \vspace{0.5cm} 
     
    \subfloat[AxelGNN$_\text{Sim}$]{%
        \includegraphics[width=\textwidth]{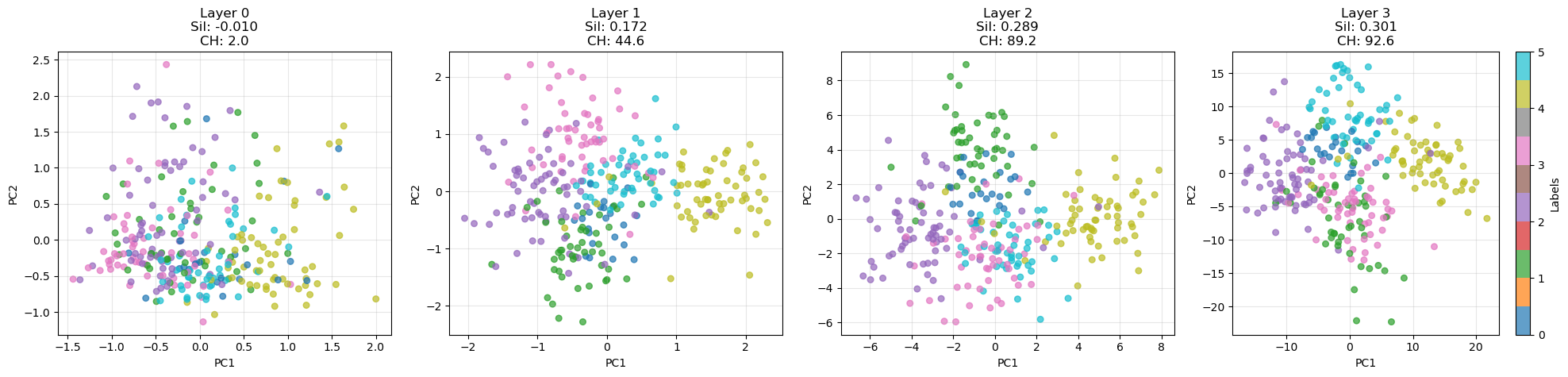}
        \label{fig:axelgnn_sim_clustering}
    }
    
    \caption{Embedding clustering evolution of AxelGNN variants across layers.}
    \label{fig:axelgnn_variants_clustering}
\end{figure}

Figure \ref{fig:axelgnn_variants_clustering} illustrates the evolution of node embeddings across network layers for both AxelGNN variants, where each subplot represents a t-SNE visualisation \cite{maaten2008visualizing} of the embeddings at different depths on the Citeseer dataset for node classification. The quality of clustering is evaluated using two metrics: the Silhouette (Sil) score \cite{shahapure2020cluster}, which measures the cohesion and separation of clusters (with higher values indicating better-defined clusters), and the Calinski-Harabasz (CH) score \cite{wang2019improved}, which assesses the ratio of between-cluster dispersion to within-cluster dispersion (with higher values indicating more distinct clustering).

Both AxelGNN variants demonstrate progressive clustering improvement across network layers. The clustering visualizations reveal distinct behaviors between AxelGNN variants. AxelGNN achieves superior clustering performance, demonstrating well-separated cluster boundaries and aggressive feature transformation. In contrast, AxelGNN$_\text{Sim}$ shows more conservative clustering behaviour, producing more compact, densely-packed clusters with reduced inter-cluster distances.

\section{Conclusion, Limitations, and Future Work}

In this work, we introduce a novel graph neural network architecture called AxelGNN that addresses fundamental limitations in traditional GNNs by incorporating principles from Axelrod's cultural dissemination model. Our approach successfully tackles three critical challenges: feature oversmoothing, inability to handle heterogeneous relationships, and monolithic feature treatment. AxelGNN adaptively handles both homophilic and heterophilic relationships within a unified framework while maintaining global polarisation to prevent oversmoothing. The proposed trait copying mechanism breaks away from monolithic feature aggregation, allowing fine-grained feature-level interactions that enhance model expressiveness. Our comprehensive experimental evaluation across node classification and 
influence estimation tasks demonstrates that AxelGNN achieves competitive or 
superior performance compared to existing GNN methods across diverse datasets 
with varying homophily-heterophily characteristics. The success of AxelGNN indicates that concepts from computational social science, particularly cultural dissemination dynamics, can offer valuable inductive biases for designing neural architectures.

 Currently, the segment size of AxelGNN is selected as a hyperparameter. While we have provided a parameter sensitivity analysis to determine a feasible value range for this, we plan to learn this value in a data-driven manner. Further, the current approach uses uniform segment sizes across all feature dimensions. Investigating adaptive segmentation strategies that dynamically determine optimal groupings based on feature correlation patterns could further enhance model performance. Beyond these architectural improvements, we plan to evaluate our method on more challenging learning scenarios such as imbalanced classification, out-of-distribution generation, adversarial attacks, and transfer learning.

\section*{Data Availability Statement}

All data supporting the findings of this study are publicly available and have been properly cited within the article.

\bibliographystyle{tfq}  
\bibliography{references}

\end{document}